\documentclass[11pt,a4paper]{article}
\usepackage[hyperref]{emnlp-ijcnlp-2019}
\usepackage{times}
\usepackage{latexsym}

\usepackage{url}

\usepackage{enumerate}
\usepackage{graphics}
\usepackage{graphicx}
\usepackage{xspace}
\usepackage{amsmath}
\usepackage{booktabs}
\usepackage{arydshln}
\usepackage{subcaption}
\usepackage{amssymb}
\graphicspath{{./figs/}}
\usepackage[compact]{titlesec}

\newcommand{\V}[1][\mathbf]{#1}

\newcommand{\bert}{\textsc{bert}\xspace}
\newcommand{\sep}{{\tt{[SEP]}}\xspace}
\newcommand{\cls}{{\tt{[CLS]}}\xspace}
\newcommand{\crf}{\textsc{crf}\xspace}
\newcommand{\sota}{\textsc{sota}\xspace}
\newcommand{\ssc}{\textsc{ssc}\xspace}
\newcommand{\csabs}{\textsc{CSAbstruct}\xspace}
\newcommand{\pubmedrct}{\textsc{PubMed-rct}\xspace}
\newcommand{\transformer}{\textsc{Transformer}\xspace}
\titlespacing\section{0pt}{12pt plus 4pt minus 2pt}{6pt plus 2pt minus 2pt}

\aclfinalcopy %

\title{Pretrained Language Models for Sequential Sentence Classification}

\DeclareSymbolFont{extraup}{U}{zavm}{m}{n}
\DeclareMathSymbol{\varheart}{\mathalpha}{extraup}{85}
\DeclareMathSymbol{\vardiamond}{\mathalpha}{extraup}{88}

\newcommand{\ai}{ $^1$}
\newcommand{\uw}{$^2$}

\makeatletter
\renewcommand*{\@fnsymbol}[1]{\ensuremath{\ifcase#1\or \bigstar \or \dagger\or \ddagger\or
    \mathsection\or \mathparagraph\or \|\or **\or \dagger\dagger
    \or \ddagger\ddagger \else\@ctrerr\fi}}
\makeatother

\author{Arman Cohan\ai \Thanks{\enspace Equal contribution.} 
\quad Iz Beltagy\ai \textcolor{darkblue}{\footnotemark[1]}  \
\quad Daniel King\ai
\quad Bhavana Dalvi\ai
\quad Daniel S. Weld\ai$^,$\uw \vspace{6pt}\\
  \ai Allen Institute for Artificial Intelligence, Seattle, WA \vspace{2pt}\\
  \uw Allen School of Computer Science \& Engineering, University of Washington, Seattle, WA \vspace{4pt}\\
  \tt{\{armanc,beltagy,daniel,bhavanad,danw\}@allenai.org}
  }

\date{}
\begin{document}
\maketitle
\begin{abstract}
    As a step toward better document-level understanding, we explore classification of a sequence of sentences into their corresponding categories, a task that requires understanding sentences in context of the document.
    Recent successful models for this task have used hierarchical models to contextualize sentence representations, and Conditional Random Fields (\crf{s}) to incorporate dependencies between subsequent labels.
    In this work, we show that pretrained language models, \bert \cite{Devlin2018BERTPO} in particular,
    can be used for this task to capture contextual dependencies without the need for hierarchical encoding nor a \crf. 
    Specifically, we construct a joint sentence representation that allows \bert Transformer layers to directly utilize contextual information from all words in all sentences.
    Our approach achieves state-of-the-art results on four datasets, including a new dataset of structured scientific abstracts.
\end{abstract}

\section{Introduction}
Inspired by the importance of document-level natural language understanding, we explore classification of a sequence of sentences into their respective roles or functions.  For example, one might classify sentences of scientific abstracts according to rhetorical roles (e.g., Introduction, Method, Result, Conclusion, etc). We refer to this task as Sequential Sentence Classification (\ssc), because the meaning of a sentence in a document is often informed by context from neighboring sentences.

Recently, there have been a surge of  new models for contextualized language representation, resulting in substantial improvements on many natural language processing tasks.
These models use multiple layers of LSTMs \cite{Hochreiter1997LongSM} or Transformers \cite{Vaswani2017AttentionIA}, and are pretrained on unsupervised text with language modeling objectives such as next word prediction \cite{Peters2018DeepCW,radford2018improving} or masked token prediction \cite{Devlin2018BERTPO,Dong2019UnifiedLM}. \bert is among the most successful models for many token- and sentence-level tasks \cite{Devlin2018BERTPO,liu2019roberta}. In addition to a masked token objective, \bert optimizes for next sentence prediction, allowing it to capture sentential context.

These objectives allow \bert to learn some document-level context through pretraining. In this work we explore the use of \bert for \ssc. For this task, prior models are primarily based on hierarchical encoders over both words and sentences, often using a Conditional Random Field (\crf) \cite{lafferty2001conditional} layer to capture document-level context \cite{Cheng2016NeuralSB,Jin2018HierarchicalNN,Chang2019LanguageMP}.
These models encode and contextualize sentences in two consecutive steps. In contrast, we propose an input representation which allows the Transformer layers in \bert to directly leverage contextualized representations of all words in all sentences, while still utilizing the pretrained weights from \bert.
Specifically, we represent all the sentences in the document as one long sequence of words with special delimiter tokens in between them.
We use the contextualized representations of the delimiter tokens to classify each sentence. The transformer layers allow the model to finetune the weights of these special tokens to encode contextual information necessary for correctly classifying sentences in context.

We apply our model to two instances of the \ssc task in
scientific text that can benefit from better contextualized representations of sentences: 
scientific abstract sentence classification and extractive summarization of scientific documents.

Our contributions are as follows:

\textit{(i)} We present
a \bert-based approach for \ssc that jointly encodes all sentences in the sequence, allowing the model to better utilize document-level context.
\textit{(ii)} We introduce and release \csabs, a new dataset of manually 
annotated sentences from computer science abstracts. Unlike biomedical abstracts which are written with explicit structure, computer science abstracts are free-form and exhibit a variety of writing styles, making our dataset more challenging than existing datasets for this task.
\textit{(iii)}  We achieve state-of-the-art (\sota) results on multiple datasets of two \ssc tasks: scientific abstract sentence classification and extractive summarization of scientific documents.\footnote{Code \& data: \url{https://github.com/allenai/sequential_sentence_classification}}

\section{Model}
\label{sec:model}

In Sequential Sentence Classification (\ssc), the goal is to 
classify each sentence in a sequence of $n$ sentences in a document.
We propose an approach for \ssc based on \bert to encode sentences in context.
The \bert model architecture consists of multiple layers of Transformer{s} and uses a specific input representation, with two special tokens, {\cls} and
{\sep}, added at the beginning of the input sentence pair and between the sentences (or bag of sentences) respectively. 
The pretrained multi-layer \transformer architecture allows the \bert model to contextualize the input over the entire sequence, allowing it to capture necessary information for correct classification. To utilize this for the \ssc task, we propose a special input representation without any additional complex architecture augmentation. Our approach allows the model to better incorporate context from all surrounding sentences.

\begin{figure}
    \centering
    \includegraphics[width=0.99\linewidth]{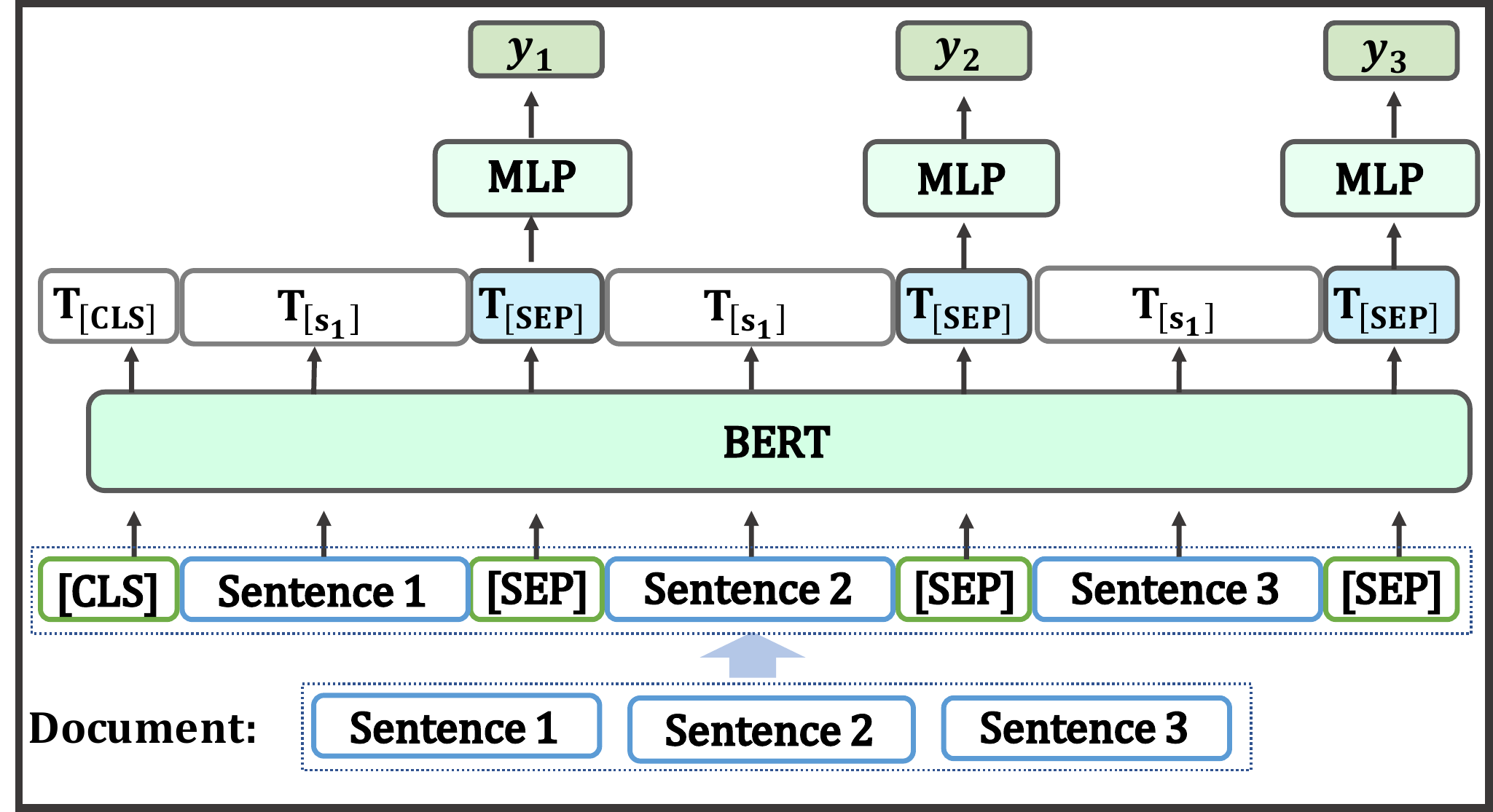}
    \caption{Overview of our model. Each {\sep} token
    is mapped to a contextualized representation
    of its sentence and then used to 
    predict a label $y_i$ for sentence$_i$.}
    \label{fig:overview}
\end{figure}

Figure~\ref{fig:overview} gives an overview of our model. Given the sequence of sentences $\V{S}=\langle\V{S_1}, ..., \V{S_n}\rangle$ we concatenate the first sentence
with \bert's delimiter, \sep, and repeat this process for each sentence, forming a large sequence containing all tokens from all sentences. After inserting the standard \cls token at the beginning of this sequence, we feed it into \bert. Unlike \bert, which uses the \cls token for classification, we use the encodings of the \sep tokens 
to classify each sentence.
We use a multi-layer feedforward network on top of the \sep representations of each sentence to classify them to their corresponding categories.\footnote{It is also possible to add another special token (e.g., \cls) at the beginning of each sentence and perform classification on that token. Empirically, we found the approaches to perform similarly.}  
Intuitively, through \bert's pretraining, the \sep tokens learn sentence structure and relations between continuous sentences (through the next sentence objective). 
The model is then finetuned on task-specific training data, 
where most of the model parameters are already pretrained using \bert and only a thin task-specific network on top is needed. During finetuning\footnote{
Following terminology from \citet{Howard2018UniversalLM},
``finetuning'' refers to ``training'' a model that was 
 previously pretrained. We use both terms interchangeably. } the model learns appropriate weights for the \sep token to allow it to capture contextual information for classifying sentences in the sequence.
This way of representing a sequence of sentences allows the self-attention layers of \bert to directly leverage contextual information from all words in all sentences, while still utilizing the pretrained weights from \bert. This is in contrast to existing hierarchical models which encode then contextualize sentences in two consecutive steps.\footnote{It is possible to add a CRF layer or another contextualizing layer on top of \sep tokens in our model, but empirically, we did not find this addition to be helpful. One explanation is that the self-attention layers of our model are already capturing necessary contextual information from the document.}

\paragraph{Handling long sequences}

Released \bert pretrained weights support sequences of up to 512
wordpieces~\cite{Wu2016GooglesNM}. This is limiting for our model on datasets where the length of each document is large, as we represent all sentences in one single sequence. However, the semantics of a sentence are usually more dependent on local context, rather than all sentences in a long document. Therefore, we set a threshold on the number of sentences in each sequence. 
We recursively bisect the document until each split has less sentences than the specified threshold. At a limit of 10 sentences, only one division is needed to fit nearly all examples for the abstract sentence classification datasets. 
A limitation of this approach is that sentences on the edge of the splits could lose context from the previous(next) split. We leave this limitation to future work.

\section{Tasks and Datasets}
\label{sec:tasks}

\begin{table}[tb]
\small
\centering
\begin{tabular}{@{}lrrrr@{}}
\toprule
 & PubMed & NICTA & CSAbstruct & CSPubSum \\ \midrule
\# docs & 20K & 1K & 2.2K & 21K \\
\# sents & 225K & 21K & 15K & 601K \\ \bottomrule
\end{tabular}
\caption{Statistics of the evaluation datasets. The first three datasets are for the abstract sentence classification task and the last dataset is for summarization.}
\label{tab:data}
\end{table}

This section describes our tasks and datasets, and any model changes that are task-specific (see Table~\ref{tab:data} for comparison of evaluation datasets).

\subsection{Scientific abstract sentence classification}
\label{sec:csdata}

This task requires classifying sentences in scientific abstracts into their rhetorical roles (e.g., \textsc{Introduction}, \textsc{Method}, \textsc{Results}, etc). We use the following three datasets in our experiments.

\paragraph{\pubmedrct}~\cite{Dernoncourt2017PubMed2R}
contains 20K biomedical abstracts from PubMed,
with sentences classified as one of 5 categories \{\textsc{Background}, \textsc{Objective}, \textsc{Method}, \textsc{Result}, \textsc{Conclusion}\}.
We use the preprocessed version of this dataset by \citet{Jin2018HierarchicalNN}.

\paragraph{\csabs}
 is a new dataset that we introduce. It has 2,189 manually annotated computer science abstracts with sentences annotated according to their rhetorical roles in the abstract, similar to the \pubmedrct categories. See \S\ref{subsec:data-construction} for details.

\paragraph{\textsc{NICTA}}~\cite{Kim2011AutomaticCO}
contains 1,000 biomedical abstracts with sentences classified into PICO categories (Population, Intervention, Comparison, Outcome)~\cite{Richardson1995TheWC}.

\subsection{Extractive summarization of scientific documents}
This task is to select a few text spans in a document
that best summarize it.
When the spans are sentences, this task can be
viewed as \ssc, classifying each sentence as a good summary sentence or not.
Choosing the best summary sentences can benefit from context of surrounding sentences. We train on \textsc{CSPubSumExt}~\cite{Collins2017ASA}, an extractive summarization dataset of 10k scientific papers, with sentences
scored as good/bad summary sentences using \textsc{Rouge} overlap
scores with paper highlights.
For evaluation, a separate test set, \textsc{CSPubSum}, of 150 publications and their paper highlights is used.\footnote{Dataset generated using author provided scripts: \url{https://github.com/EdCo95/scientific-paper-summarisation}}

A key difference between the training of our model and that of \citet{Collins2017ASA} is that they use the \textsc{Rouge} scores to label the top (bottom) 20 sentences
 as positive (negative), and the rest
are neutral. However, we found it better to train our
model to directly predict the \textsc{Rouge}
scores, and the loss function we used is Mean Square Error.

\setlength{\dashlinedash}{0.2pt}
\setlength{\dashlinegap}{1.5pt}
\setlength{\arrayrulewidth}{0.4pt}

\begin{table}[]
\small
\centering
\begin{tabular}{@{}lrrrr@{}}
\toprule
 \multicolumn{5}{c}{CSAbstruct characteristics} \\ \midrule
Doc length (sentences) & \multicolumn{2}{l}{avg~:~ ~~6.7} & \multicolumn{2}{r}{std~:~ 1.99} \\ \hdashline
Sentence length (words) & \multicolumn{2}{l}{avg~:~ 21.8} & \multicolumn{2}{r}{std~:~ 10.0} \\ \hdashline
Label distribution & \multicolumn{2}{r}{\begin{tabular}[c]{@{}l@{}}\textsc{Background}\\ \textsc{Method}\\ \textsc{Result}\\ \textsc{Objective}\\ \textsc{Other}\end{tabular}} & \multicolumn{2}{r}{\begin{tabular}[c]{@{}l@{}}0.33\\ 0.32\\ 0.21\\ 0.12\\ 0.03\end{tabular}} \\ \bottomrule
\end{tabular}
\caption{Characteristics of our \csabs dataset}
\label{tab:csabs}
\end{table}

\subsection{\csabs construction details}
\label{subsec:data-construction}

\csabs is a new dataset of annotated computer science abstracts with sentence labels according to their rhetorical roles. The key difference between this dataset and \pubmedrct is that PubMed abstracts are written according to a predefined structure, whereas computer science papers are free-form. Therefore, there is more variety in writing styles in \csabs. \csabs is collected from the Semantic Scholar corpus \cite{Ammar2018ConstructionOT}.
Each sentence is annotated by 5 workers on the Figure-eight platform,\footnote{\url{http://figure-eight.com}} with one of 5 categories \{\textsc{Background}, \textsc{Objective}, \textsc{Method}, \textsc{Result}, \textsc{Other}\}. Table \ref{tab:csabs} shows characteristics of the dataset. We use 8 abstracts (with 51 sentences) as test questions to train crowdworkers. Annotators whose accuracy is less than 75\% are disqualified from doing the actual annotation job. The annotations are aggregated using the agreement on a single sentence weighted by the accuracy of the annotator on the initial test questions. A confidence score is associated with each instance based on the annotator initial accuracy and agreement of all annotators on that instance. We then split the dataset 75\%/15\%/10\% into train/dev/test partitions,
 such that the test set has the highest confidence scores.
Agreement rate on a random subset of 200 sentences is 75\%, which is quite high given the difficulty of the task. Compared with \pubmedrct, our dataset exhibits a wider variety of writing styles, since its abstracts are not written with an explicit structural template.

\setlength{\dashlinedash}{0.2pt}
\setlength{\dashlinegap}{1.5pt}
\setlength{\arrayrulewidth}{0.4pt}

\begin{table}[t]
\centering
\setlength{\tabcolsep}{2pt}
\small
\begin{tabular}{@{}lrrr@{}}
\toprule
Model & \textsc{PubMed} & \textsc{CSAbst.}  & \textsc{NICTA} \\ \midrule
\citet{Jin2018HierarchicalNN} & 92.6 & 81.3 & 84.7 \\ \hdashline
\bert+Transformer & 89.6 & 78.8 & 78.4 \\
\bert+Transformer+\crf & 92.1 & 78.5   & 79.1 \\ \hdashline
Our model & \bf{92.9} & \bf{83.1}  & \bf{84.8} \\ \bottomrule
\end{tabular}
\caption{Abstract sentence classification (micro F1).}
\label{tab:results}
\end{table}

\section{Experiments}
\paragraph{Training and Implementation}
We implement our models using AllenNLP \cite{Gardner2017AllenNLP}. We use \textsc{SciBERT} pretrained weights \cite{Beltagy2019SciBERT} in both our model and \bert-based baselines, because our datasets are from the scientific domain.
As in prior work~\cite{Devlin2018BERTPO,Howard2018UniversalLM},
for training, we use dropout of 0.1, the Adam~\cite{Kingma2015AdamAM}
optimizer for 2-5 epochs,
and learning rates of $5e\text{-}6$, $1e\text{-}5$, $2e\text{-}5$, or $5e\text{-}5$.
We use the largest batch size that fits in the memory of a Titan~V GPU
(between 1 to 4 depending on the dataset/model)
and use gradient accumulation for effective batch size of 32.
We report the average of results from 3 runs with different random seeds for the abstract sentence classification datasets to control potential non-determinism associated with deep neural models \cite{Reimers2017ReportingSD}. For summarization, we use the best model on the validation set. We choose hyperparameters based on the best performance on the validation set. We release our code and data to facilitate reproducibility.\footnote{\small{\url{https://github.com/allenai/sequential_sentence_classification}}}

\paragraph{Baselines}
We compare our approach with two strong \bert-based baselines, finetuned for the task.
The first baseline, {\bert}+Transformer, uses the \cls token
to encode individual sentences as described in \citet{Devlin2018BERTPO}. We add an additional Transformer layer over
the \cls vectors to contextualize the sentence representations over the entire sequence.
The second baseline, {\bert}+Transformer+\crf, additionally adds a \crf layer.
Both baselines split long lists of sentences into splits of length 30 using the method in \S\ref{sec:model} to fit into the GPU memory.

We also compare with existing SOTA models for each dataset.
For the \pubmedrct and \textsc{NICTA} datasets, we report the results of  \citet{Jin2018HierarchicalNN}, who use a hierarchical LSTM model augmented with attention and CRF.
We also apply their model on our dataset, \csabs,
using the authors' original implementation.\footnote{\small{\url{https://github.com/jind11/HSLN-Joint-Sentence-Classification}}}
For extractive summarization, we compare to \citet{Collins2017ASA}'s model, \texttt{SAF+F Ens}, the model with highest reported results on this dataset. This model is an ensemble of an LSTM-based model augmented with global context and abstract similarity features, and a model trained on a set of hand-engineered features.

\begin{table}[t]
\centering
\small
\setlength{\tabcolsep}{2pt}
\begin{tabular}{@{}lrrr@{}} \toprule
Model & \textsc{Rouge-L} \\ \midrule
SAF + F Ens \cite{Collins2017ASA}  & 0.313  \\ \hdashline
\bert+Transformer & 0.287 \\ \hdashline
Our model & 0.306 \\
Our model + \textsc{AbstractRouge} & \textbf{ 0.314} \\ \bottomrule

\end{tabular}
\caption{Results on \textsc{CSPubSum} }
\label{tab:summarization}
\end{table}

\subsection{Results}
Table \ref{tab:results} summarizes results for abstract sentence
classification. Our approach achieves state-of-the-art results
on all three datasets, outperforming~\citet{Jin2018HierarchicalNN}.
It also outperforms our {\bert}-based baselines. The performance gap between our baselines and our best model is large for small datasets
(\csabs, \textsc{NICTA}), and smaller
for the large dataset (\pubmedrct). This suggests
the importance of pretraining for small datasets.

Table~\ref{tab:summarization} summarizes results on \textsc{CSPubSum}. Following \citet{Collins2017ASA} we take the top 10 predicted sentences as the summary and use \textsc{Rouge-L} scores for evaluation.
It is clear that our approach outperforms {\bert}+\textsc{Transformer}.
The \bert+\textsc{Transformer}+\crf baseline is not included here because,
as mentioned in section~\ref{sec:tasks}, we train our model to predict \textsc{Rouge}, not binary labels as in~\citet{Collins2017ASA}.
As in~\citet{Collins2017ASA}, we found the \textsc{Abstract-rouge}
feature to be useful. Our model augmented with this feature slightly outperforms ~\citet{Collins2017ASA}'s model, which is a relatively complex ensemble model and
uses a number of carefully engineered features for the task. Our model is a single model with only one added feature.

\paragraph{Analysis}

\begin{figure}[t]
\begin{subfigure}[b]{0.49\linewidth}
\includegraphics[width=0.95\linewidth]{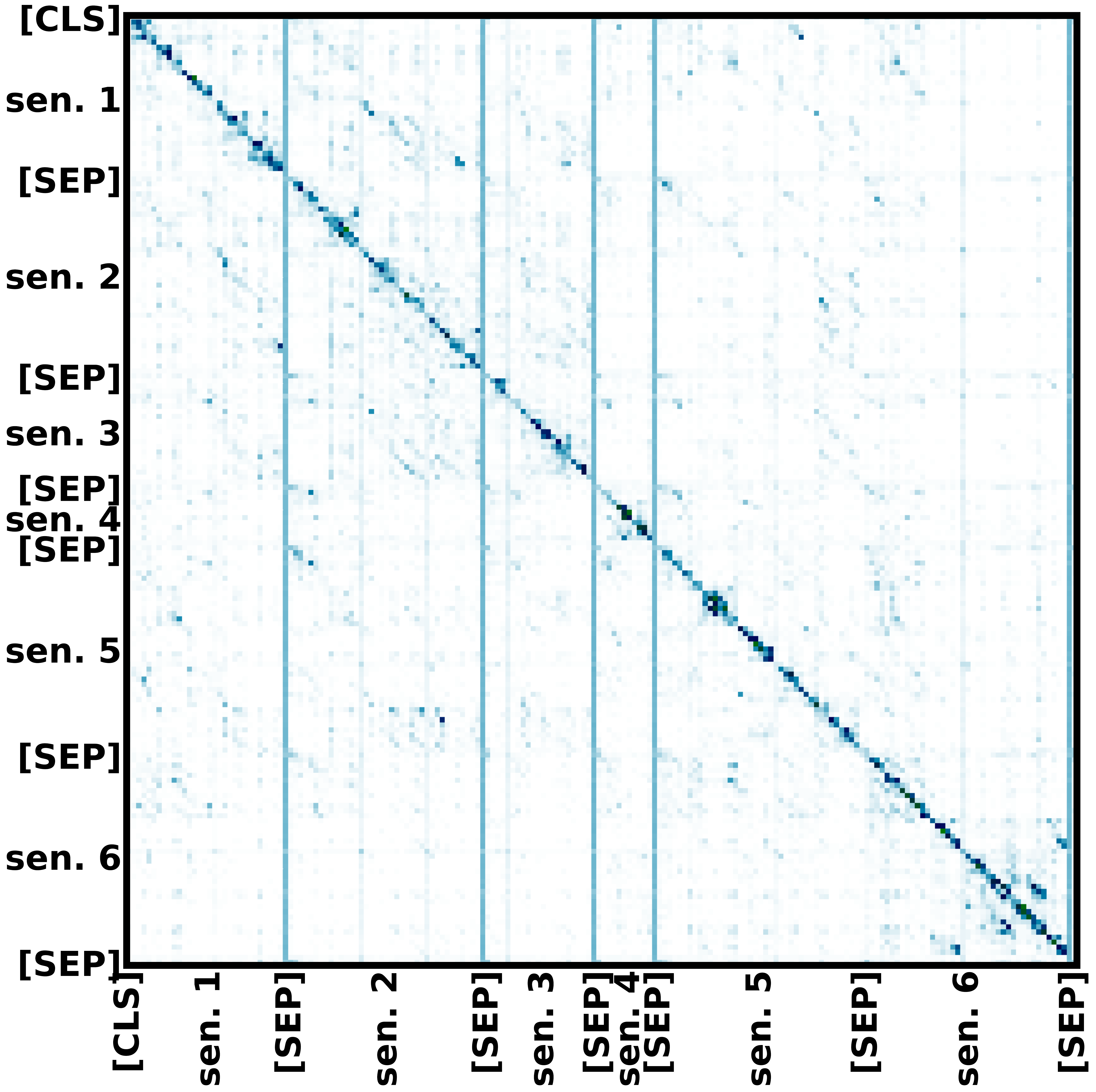}
\caption{Before finetuning}
\label{fig:attn_fine_frozen}
\end{subfigure}
\begin{subfigure}[b]{0.49\linewidth}
\includegraphics[width=0.95\linewidth]{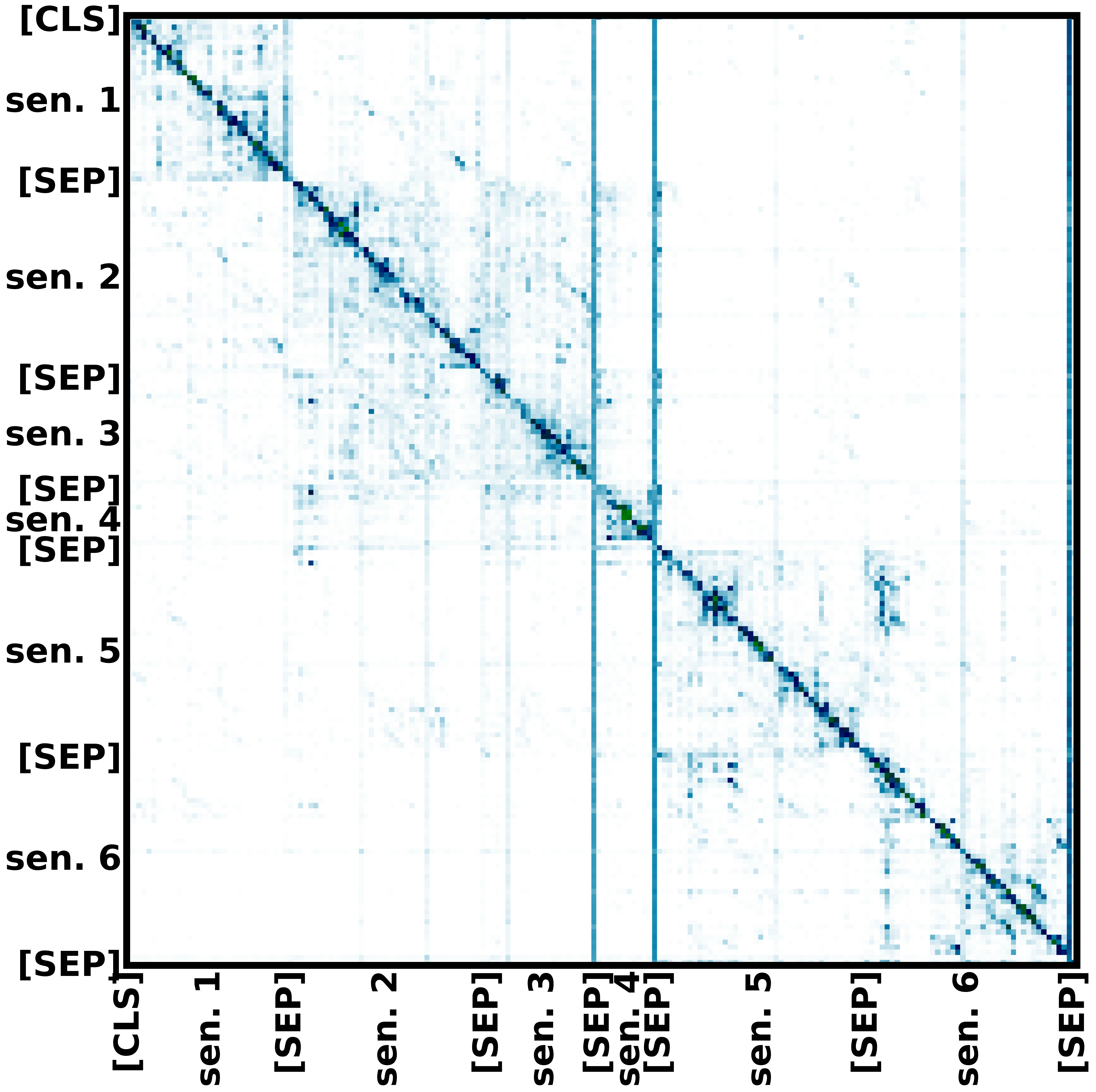}
\caption{After finetuning}
\label{fig:attn_fine_finetune}
\end{subfigure}
\caption{
Self-attention weights of the top 2 layers of \bert for one abstract. Cell value in row i, column j, is the maximum attention weight of token i attending to token j across all 12 Transformer attention heads.}
\label{fig:attn_fine}
\vspace{-10pt}
\end{figure}

To better understand the advantage of our joint sentence encoding relative to the {\bert}+Transformer baseline, we qualitatively analyze examples from \csabs that our model gets right and the baseline gets wrong. We found that 34/134 of such examples require context to classify correctly.\footnote{Of the 1349 examples in the test set, our model gets 134 correct that the \bert+Transformer baseline gets wrong, and the baseline gets 79 correct that our model gets wrong.}

For example, sentences 2 and 3 from one abstract are as follows: ``\textit{We present an improved oracle for the arc-eager transition system, which provides a set of optimal transitions [...]}.'', ``\textit{In such cases, the oracle provides transitions that will lead to the best reachable tree [...].}''. In isolation, the label for sentence 3 is ambiguous, but with context from the previous sentence, it clearly falls under the \textsc{Method} category. 

Figure~\ref{fig:attn_fine} shows \bert self-attention weights for the above-mentioned abstract
before and after finetuning. Before (Figure~\ref{fig:attn_fine_frozen}),
attention weights don't exhibit a clear pattern.
After (Figure~\ref{fig:attn_fine_finetune}),
we observe blocks along the matrix diagonal of
sentences attending to themselves, except for the block encompassing sentences 2 and 3.
The words in these two sentences attend to each other, 
enabling the encoding of sentence 3 to capture the information needed from
sentence 2 to predict its label (see Appendix~\ref{sec:abs_appendix} for additional patterns).

\section{Related Work}

Prior work on scientific Sequential Sentence Classification datasets (e.g. \pubmedrct and \textsc{NICTA}) use hierarchical sequence encoders (e.g. LSTMs) to encode each sentence and contextualize the encodings, and apply \crf on top \cite{Dernoncourt2017PubMed2R, Jin2018HierarchicalNN}. 
Hierarchical models are also used for summarization~\cite{Cheng2016NeuralSB,Nallapati2016SummaRuNNerAR,Narayan2018RankingSF}, usually trained in a seq2seq fashion \cite{sutskever2014sequence} and evaluated on newswire data such as the CNN/Daily mail benchmark~\cite{hermann2015teaching}. Prior work proposed generating summaries of scientific text by leveraging citations~\cite{Cohan2015ScientificAS}
and highlights~\cite{Collins2017ASA}. 
The highlights-based summarization dataset introduced by \citet{Collins2017ASA} is among the largest 
extractive scientific summarization datasets.
Prior work focuses on specific architectures designed for each of the tasks described in \S\ref{sec:tasks}, giving them more power to model each task directly. Our approach is more general, uses minimal architecture augmentation, leverages language model pretraining, and can handle a variety of \ssc tasks.

\section{Conclusion and Future Work}
\vspace{6pt}
We demonstrated how we can leverage pretrained language models, in particular \bert, for \ssc without additional complex architectures. We showed that jointly encoding sentences in a sequence results in improvements across multiple datasets and tasks in the scientific domain.
For future work, 
we would like to explore methods for better encoding long sequences using pretrained language models. 

\section*{Acknowledgments}
\vspace{6pt}
 We would like to thank Matthew Peters, Waleed Ammar and Hanna Hajishirzi for helpful discussions, Madeleine van Zuylen for help in crowdsourcing and data analysis, and the three anonymous reviewers for their comments and suggestions. 
Computations on \url{beaker.org} were supported in part by credits
from Google Cloud. Other support includes ONR grant N00014-18-1-2193 and the WRF/Cable Professorship, 

\bibliography{emnlp-ijcnlp-2019}
\bibliographystyle{acl_natbib}

\newpage
\appendix
\section{Additional analysis}
\label{sec:abs_appendix}

\begin{figure}[h]
\centering
\begin{subfigure}[b]{0.49\linewidth}
\centering
\includegraphics[width=0.83\linewidth]{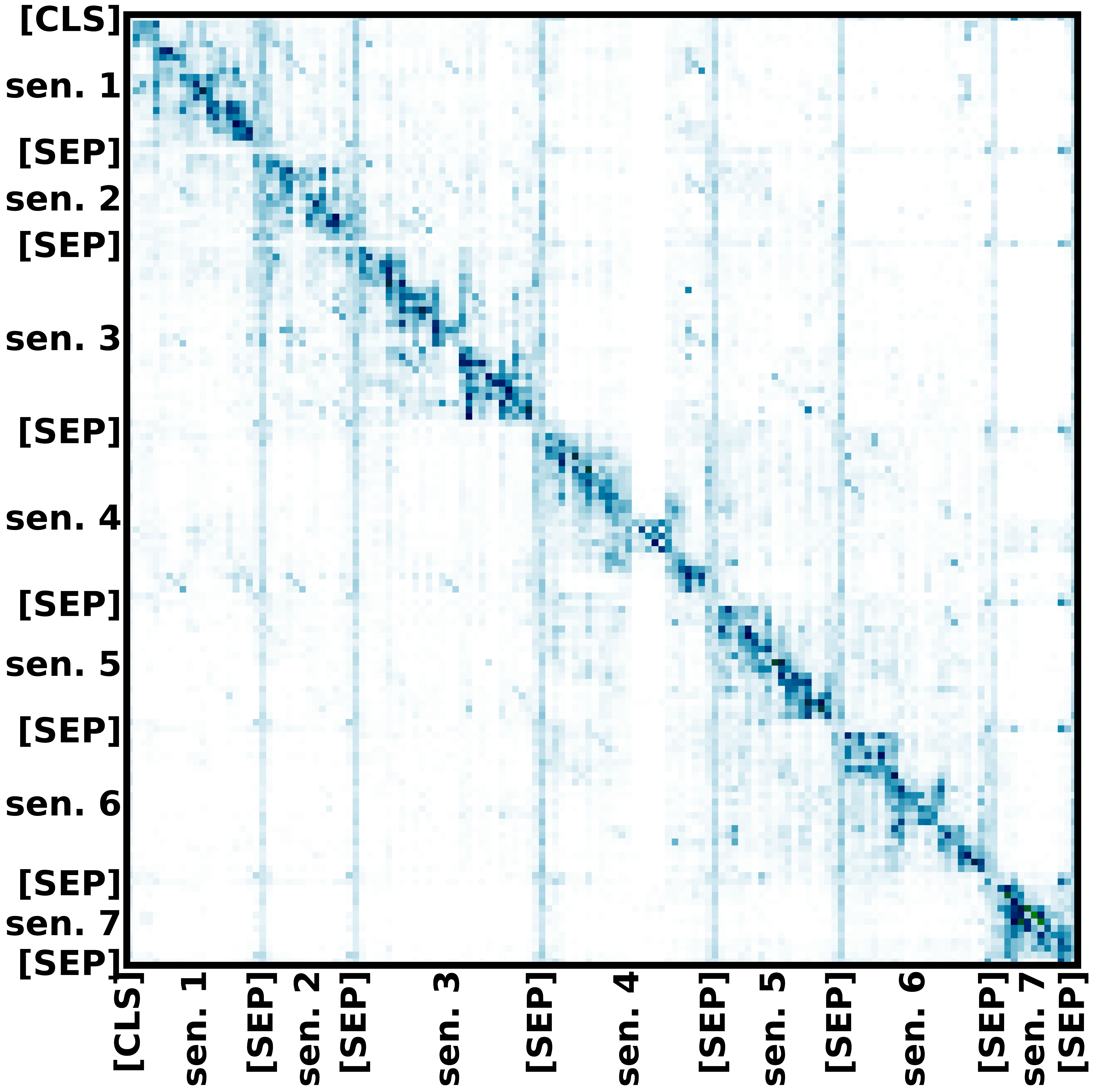}
\caption{Before finetuning}
\end{subfigure}
\begin{subfigure}[b]{0.49\linewidth}
\centering
\includegraphics[width=0.83\linewidth]{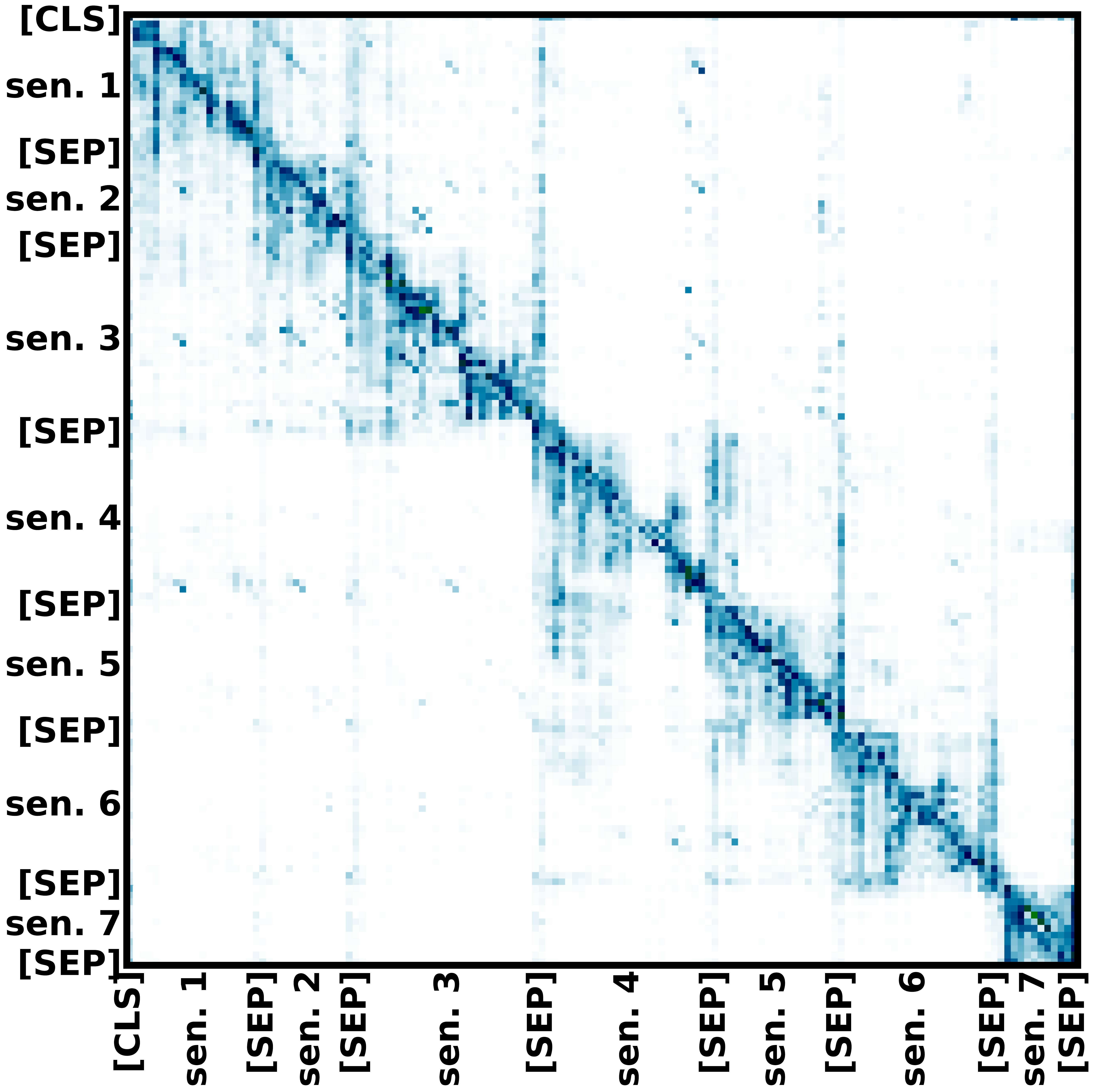}
\caption{After finetuning}
\end{subfigure}
\caption{Visualization of attention weights for layer 8 of \bert before and after finetuning.}
\label{fig:analysis-layer8}
\end{figure}

\begin{figure}[h]
\centering
\begin{subfigure}[b]{0.49\linewidth}
\centering
\includegraphics[width=0.83\linewidth]{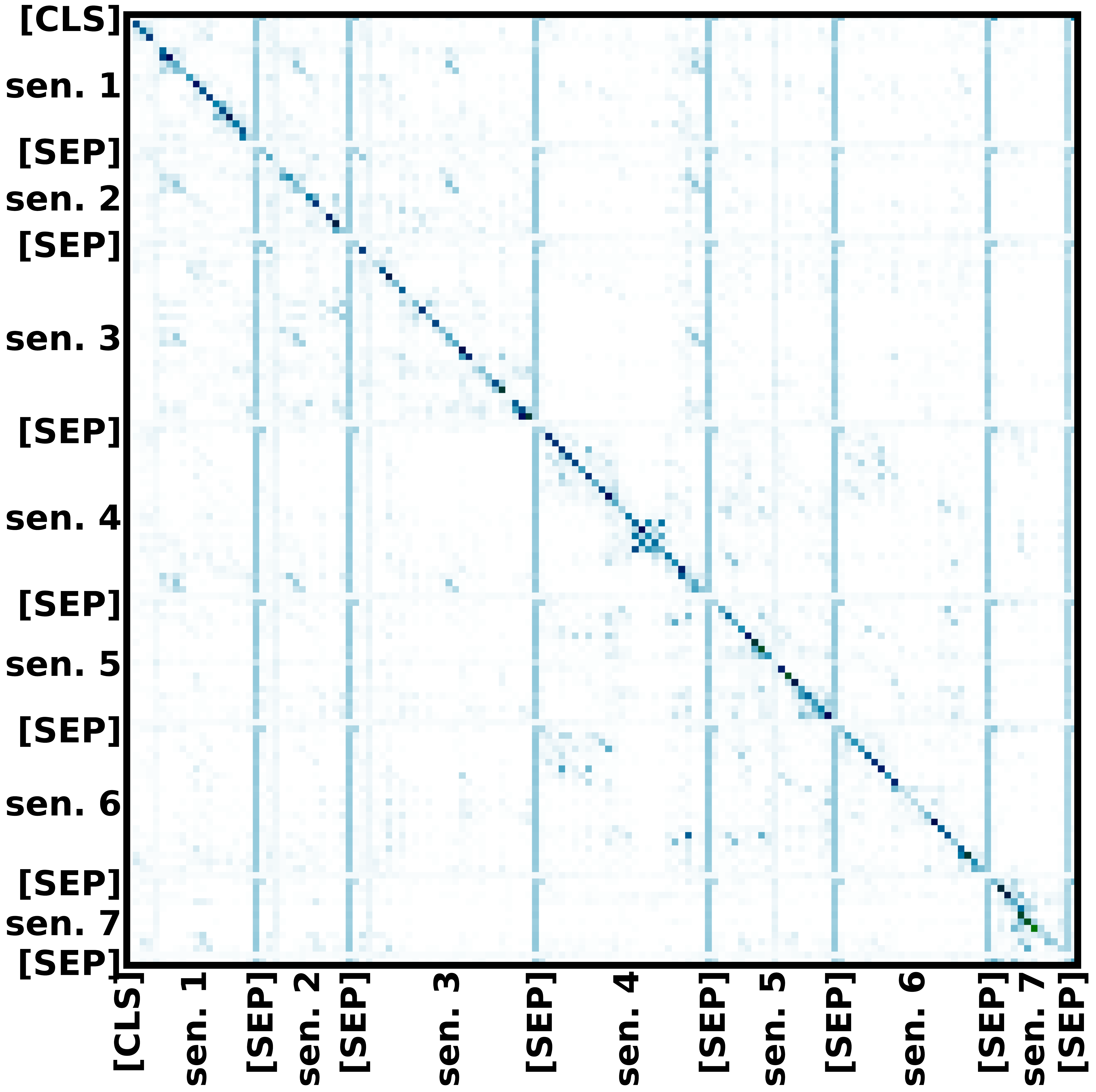}
\caption{Before finetuning}
\end{subfigure}
\begin{subfigure}[b]{0.49\linewidth}
\centering
\includegraphics[width=0.83\linewidth]{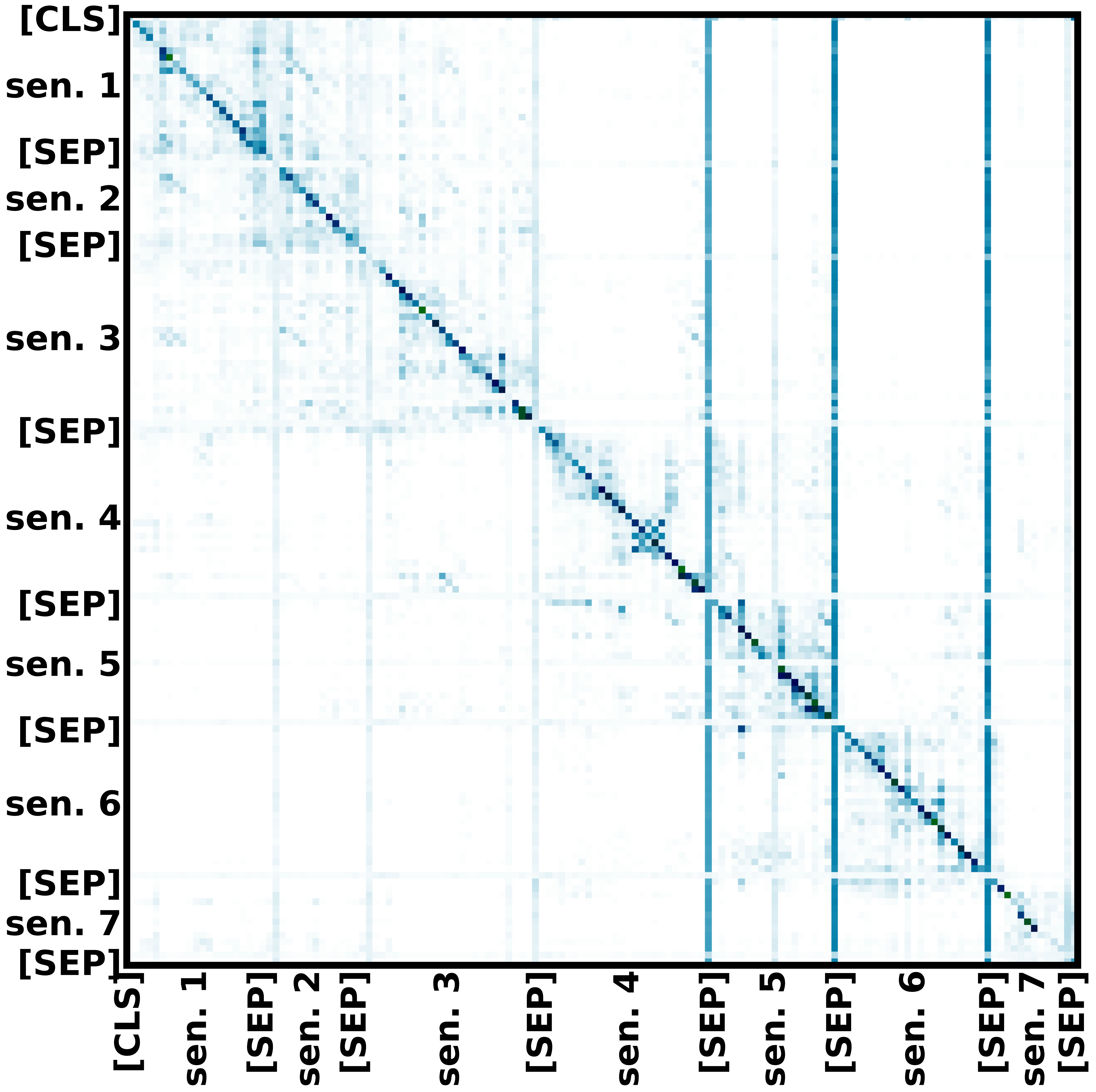}
\caption{After finetuning}
\end{subfigure}
\caption{Visualization of attention weights in final layer (layer 12) of \bert before and after finetuning.}
\label{fig:analysis-layer12}
\end{figure}

Figures \ref{fig:analysis-layer8} and \ref{fig:analysis-layer12} show attention weights of \bert before and after finetuning. We observe that before finetuning, the attention patterns on \sep tokens and periods is almost identical between sentences. However, after finetuning, the model attends to sentences differently, likely based on their different role in the sentence that requires different contextual information.

\end{document}